\documentclass[10pt,twocolumn,letterpaper]{article}

\usepackage[pagenumbers]{iccv} % To force page numbers, e.g. for an arXiv version
\usepackage[accsupp]{axessibility}  % Improves PDF readability for those with disabilities.
\definecolor{iccvblue}{rgb}{0.21,0.49,0.74}
\usepackage[pagebackref,breaklinks,colorlinks,allcolors=iccvblue]{hyperref}
\usepackage{multicol}
\usepackage{multirow}
\usepackage{graphicx}
\usepackage{animate}
\usepackage[normalem]{ulem}
\usepackage{subcaption}
\def\paperID{} % *** Enter the Paper ID here
\def\confName{ICCV}
\def\confYear{2025}

\title{On-Device Diffusion Transformer Policy for Efficient Robot Manipulation}

\author{Yiming Wu$^1$\ \hspace{12pt} Huan Wang$^2$\thanks{Corresponding authors: Huan Wang and Dong Xu.}\ \hspace{12pt} Zhenghao Chen$^3$\ \hspace{12pt} Jianxin Pang$^4$\ \hspace{12pt} Dong Xu$^1$$^*$\ \\
$^1$ School of Computing and Data Science, The University of Hong Kong \\
$^2$ School of Engineering, Westlake University \\
$^3$ School of Information and Physical Sciences, University of Newcastle\\
$^4$ UBTech Robotics Corp.\\
{\tt\small {\{yimingwu, dongxu\}@hku.hk \hspace{12pt} wanghuan@westlake.edu.cn \hspace{12pt} zhenghao.chen@newcastle.edu.au}}
}

\newcommand{\bx}{\boldsymbol{x}} % prelimary diffusion model

\newcommand{\obs}{\boldsymbol{o}} % observation
\newcommand{\act}{\boldsymbol{a}} % action
\newcommand{\T}{\mathcal{T}} % trajectory set
\newcommand{\g}{\mathbf{g}} % goal
\newcommand{\E}{\boldsymbol{E}}
\newcommand{\D}{\boldsymbol{D}}
\newcommand{\G}{\boldsymbol{G}}
\newcommand{\W}{\boldsymbol{W}}

\newcommand{\layer}{\boldsymbol{\phi}}
\newcommand{\mask}{\mathcal{M}}

\begin{document}
\maketitle
\begin{abstract}

  Diffusion Policies have significantly advanced robotic manipulation tasks via imitation learning, but their application on resource-constrained mobile platforms remains challenging due to computational inefficiency and extensive memory footprint. In this paper, we propose LightDP, a novel framework specifically designed to accelerate Diffusion Policies for real-time deployment on mobile devices. LightDP addresses the computational bottleneck through two core strategies: network compression of the denoising modules and reduction of the required sampling steps. We first conduct an extensive computational analysis on existing Diffusion Policy architectures, identifying the denoising network as the primary contributor to latency. To overcome performance degradation typically associated with conventional pruning methods, we introduce a unified pruning and retraining pipeline, optimizing the model's post-pruning recoverability explicitly. Furthermore, we combine pruning techniques with consistency distillation to effectively reduce sampling steps while maintaining action prediction accuracy. Experimental evaluations on the standard datasets, \ie, PushT, Robomimic, CALVIN, and LIBERO, demonstrate that LightDP achieves real-time action prediction on mobile devices with competitive performance, marking an important step toward practical deployment of diffusion-based policies in resource-limited environments. Extensive real-world experiments also show the proposed LightDP can achieve performance comparable to state-of-the-art Diffusion Policies.

\end{abstract}

\section{Introduction}
\label{sec:intro}

Diffusion Policies have demonstrated significant success in robotic manipulation tasks through imitation learning, as evidenced by various studies~\cite{zhao23aloha,fu2024mobile,brohan23rt1,zitkovich23rt2,chi2023diffusion,reuss24mdt,reuss2025mode,li24roboflamingo,wu2024gr1,kim2024openvla,black2024pi0,octo2024octo}. This success fuels the ambition to deploy general-purpose embodied agents in robots, particularly those with limited computation resources. However, this endeavor presents multifaceted challenges: 1) Diffusion Policies require multiple denoising steps, which slows down the generation process;  2) the standard architectures~\cite{chi2023diffusion,reuss24mdt,reuss2025mode} involve billions of parameters, leading to high memory usage. These factors impede real-time applications on resource-constrained platforms like mobile robots and drones. 
To address these challenges, recent work by DeeR-VLA~\cite{yue2024deervla} introduces a multi-exit architecture built on the Roboflamingo framework~\cite{li24roboflamingo}, enabling dynamic termination of the computation process to accelerate action prediction. While this design achieves considerable computation reduction on GPU devices, its early exit strategy remains suboptimally tuned for mobile platforms.

In this work, we introduce a novel framework named~\emph{LightDP} for Diffusion Policies that enables models to achieve real-time generation on mobile devices. 
To achieve this, we mainly focus on two primary strategies: compressing the denoising network to improve the inference speed and reducing the sampling steps. First, we provide an analysis of two Diffusion Policies named DiffusionPolicy Transformer (DP-T)~\cite{chi2023diffusion} and MDT-V~\cite{reuss24mdt}. Through the comprehensive component evaluation, we observe that the denoiser is the major bottleneck for Diffusion Policies (as shown in Table~\ref{tab:time analysis}). In this work, we follow the conventional model pruning pipeline, in which the model is pruned and re-trained to resist the performance drop. In previous pruning approaches based on importance metrics~\cite{men2024shortgpt}, oracle design~\cite{kimBKSDMLightweightFast2023}, or lottery hypothesis~\cite{frankle2018lottery}, the pruning and retraining process is separated, which can lead to suboptimal performance. In contrast, we integrate the pruning and retraining process in a unified framework, which can enhance the recoverability of the Diffusion Policies and explicitly model and optimize the post-finetuning performance of pruned models. 
Second, reducing the sampling steps is another straightforward way to speed up diffusion policies, but it would result in inevitable performance degradation without distillation. To preserve the prediction of initial action with fewer inference steps, we integrate the pruning strategies introduced with consistency distillation~\cite{song2023consistency,song2024improved}. With the proposed~\emph{LightDP}, we show efficient diffusion policies on mobile devices, which can achieve real-time generation with competitive performance in three data sets. Our contributions are summarized as follows:
\begin{itemize}
  \item We present a novel framework for Diffusion Policies to obtain the efficient diffusion transformer that achieves real-time action prediction on the mobile device significantly faster than the original models.
  \item To our knowledge, this is the first work to address deploying Diffusion Policies on mobile devices. We provide a comprehensive analysis of these policies' computational cost and memory footprint.
  \item We integrate the pruning and step distillation process in a unified framework that enhances the recoverability of the models under the extensive benchmarking on the widely used datasets, \eg,~Push-T, Robomimic, CALVIN, and LIBERO. The extensive real-world evaluations present the effectiveness of our approach in practical scenarios.
\end{itemize}

\section{Related Work}
\label{sec:related}

\subsection{Diffusion Policies}
\label{sec:related:diffusion policies}

Several studies have investigated the application of diffusion models ~\cite{song2020denoising,karras2022elucidating,ho2021classifier} on policy learning, such as BESO~\cite{reuss2023goalconditionedimitationlearningusing} Diffusion Policy~\cite{chi2023diffusion}, MDT~\cite{reuss24mdt}, and MoDE~\cite{reuss2025mode}. 
Some approaches integrate pretrained visual-language models~\cite{reuss24mdt} directly into end-to-end visuomotor manipulation policies but these often involve significant architectural constraints or require calibrated cameras, limiting their generalizability. Further extension on 3D representations~\cite{Ze2024DP3} enable the model to tackle complex 3D robotic manipulation tasks, demonstrating superior performance compared to traditional methods.
Despite the success of these methods, they often require extensive fine-tuning and are computationally expensive, limiting their deployment on resource-constrained devices. Reuss~\etal~\cite{reuss2025mode} propose an MoE-based policy network that can be trained end-to-end, and only a few parameters are activated during inference, reducing the computational cost significantly. And some concurrent work~\cite{fu2024mobile,yue2024deervla,black2025real,shukor2025smolvla} explored accelerating the inference of VLA models.

In this work, we focus on compressing the policy models and deploying the model on resource-constrained devices, such as smartphones and NVIDIA Jetson devices.

\subsection{Network Pruning for Diffusion Models}
Due to the significant computational demands of diffusion models, many works aim to enhance efficiency by either pruning network components~\cite{han2015learning,li2017pruning,wang2021neural,fang2023depgraph, chen2024group,chen2023neural} or employing knowledge distillation~\cite{hinton2015distilling,salimans2022progressive,meng2023distillation}. The former targets reducing the model's size while the latter cuts down on the number of required denoising steps. For instance, Li~\textit{et al.} introduced SnapFusion~\cite{liSnapFusionTextImageDiffusion2023}, an early method that accelerates diffusion models by modifying the architecture through channel and block pruning alongside distillation techniques.
SnapFusion determines the importance of each block by evaluating both the degradation in CLIP score and the gain in inference speed, and the blocks are removed using a ``trial-and-error'' procedure~\cite{mozer1988skeletonization,molchanov2017pruning}: those causing the \textit{smallest} drop in CLIP score and the \textit{largest} boost in speed are considered less critical. Additionally, SnapFusion incorporates a CFG-aware distillation loss to better align the outputs of a pruned (student) model with those of its original (teacher) one after classifier-free guidance is applied. 

In a similar vein, BK-SDM~\cite{kimBKSDMLightweightFast2023} accelerates Stable Diffusion by eliminating entire weight blocks, although it relies solely on the CLIP score to assess importance. A subsequent finetuning step based on feature distillation helps recover performance, achieving a reduction in model size of around 30\% to 50\% with marginal performance loss. The resultant model is then further refined into EdgeFusion~\cite{castells2024edgefusion} based on a robust distillation method named LCM~\cite{luoLATENTCONSISTENCYMODELS2023}.

Furthermore, Google's MobileDiffusion~\cite{zhaoMobileDiffusionSubsecondTextImage2023} applies pruning to shrink model size but goes a step further by introducing additional architectural modifications. These include adding more transformer layers in the U-Net's intermediate stages, reducing the number of channels, and decoupling self-attention from cross-attention to enhance performance. Complemented by a specific distillation loss inspired by SnapFusion and UFOGen~\cite{xu2024ufogen}, it achieves remarkably fast inference speeds reportedly around 0.2 seconds on iPhone 15 Pro. 

In parallel, SANA-1.5~\cite{xie2025sana1.5} presents a linear diffusion transformer that introduces a block-level importance analysis for model depth pruning, enabling compression to arbitrary sizes with minimal quality drop. The pruned SANA models can even be scaled back up at inference via a repeated sampling strategy to match larger-model performance. In the realm of on-device applications, Edge-SD-SR~\cite{hadji2025edge} adapts Stable Diffusion for super-resolution by trimming the model to only ~169M parameters through a specialized bidirectional conditioning design and joint training, enabling 4$\times$ upscaling in ~1.1s on mobile hardware while matching or surpassing dedicated super-resolution methods in quality.

\section{Preliminaries}
\label{sec:prelimary}

\textbf{Diffusion Models.}\label{sec:preliminaries:video_diffusion_models} 
Diffusion models~\cite{song2021scorebased,karras2022elucidating} are a class of generative models that iteratively produce data by gradually adding and removing noise. They involve two main processes: 
1. Forward Diffusion Process: Noise is progressively added to the input data, transforming it into a noise-like distribution. 
2. Reverse Denoising Process: The original input is reconstructed from the noisy data by progressively removing the added noise. 
Within a continuous-time framework, adding independent and identically distributed (i.i.d.) Gaussian noise with standard deviation $\sigma$ to the data distribution $p_{\text{data}}(\bx_0)$ results in a noisy distribution $p(\bx; \sigma)$.
As $\sigma$ increases from a small value $\sigma_{\min}$ to a large value $\sigma_{\max}$, $p(\bx; \sigma_{\max})$ approximates pure noise. The probability flow ordinary differential equation (PF-ODE) describes the evolution of the data under this noise addition:
\begin{equation}\label{eq:pf-ode}
  \mathrm{d}\bx = -\dot{\sigma}_t \, \sigma_t \, \nabla_{\bx} \log p(\bx, \sigma_t) \, \mathrm{d}t,
\end{equation}
where $\nabla_{\bx} \log p(\bx, \sigma_t)$ is the score function, often approximated by $\frac{D_\theta(\bx; \sigma_t) - \bx}{\sigma_t^2}$. Within the EDM~\cite{karras2022elucidating} framework, the denoising function $D_\theta(\bx_t, \sigma_t)$ is parameterized as:
\begin{equation}\label{eq:edm}
  D_{\theta}=c_{skip}(t)\bx_t+c_{out}(t)f_{\theta}(c_{in}(t)\bx_t, c_{noise}(t)),
\end{equation}
where $f_\theta$ is a neural network trained to minimize the $L^2$ denoising error, and $c_{\text{skip}}$, $c_{\text{in}}$, $c_{\text{out}}$, and $c_{\text{noise}}$ are time-dependent coefficients.

\noindent\textbf{Consistency Models.}\label{sec:preliminaries:consistency_models} Consistency models, a family of generative models, are designed to generate data efficiently by directly mapping noisy inputs to their clean counterparts in a single step. They enforce a self-consistency property that ensures the model's outputs remain invariant across different noise levels, \ie , $f_\theta(\bx_t, t) = f_\theta(\bx_{t^\prime}, t^\prime)$, where $\bx_t$ and $\bx_{t'}$ are samples taken at different time steps $t$ and $t'$ along the ODE trajectory. In the EDM framework, consistency models adopt the boundary conditions $c_{\text{skip}}(0) = 1$ and $c_{\text{out}}(0) = 0$. One approach to training these models, known as consistency distillation, involves refining a pre-trained diffusion model by minimizing the consistency loss:
\begin{equation}
  \begin{aligned}
    \mathcal{L}_{CD}( & \theta, \theta^- ; \Psi)= \\
     &\mathbb{E} \left[ d\left( f_\theta \left( \bx_{t_{n+k}}, t_{n+k},  \right), f_{\theta^-} \left( \hat{\bx}_{t_n}^{\Psi, \omega}, t_n,\right) \right) \right],
  \end{aligned}
  \label{eq:consistency distillation}
\end{equation}
where $d$ is a distance function, $\hat{\bx}_{t_n}^{\Psi, \omega}$ is the data reversed by an ODE solver $\Psi$ with classifier-free guidance weight $\omega$, $n$ is the time step of the pre-trained diffusion model, and $k$ is the step interval.

\section{Method}
\label{sec:method}

\subsection{Problem Formulation}
\label{sec:method:overview}
Recent advances in imitation learning have enabled robots to learn complex manipulation tasks from demonstrations collected by human experts. Given the demonstration $\T$, 
a trajectory $\tau \in \T$ is a sequence of observation $\obs$ and robot action $\act$, denoted as $\tau = \{(\obs_1, \act_1), ..., (\obs_{N_\tau}, \act_{N_\tau})\}$. 
A diffusion policy $\pi_\phi({\act | \obs, \g})$ is trained to imitate the expert's behavior by maximizing the log-likelihood of the action $\act$ given the observation $\obs$ and goal $\g$. 
Under the multi-modal setting, the goal $\g$ is a high-level instruction that specifies the desired outcome of the task, could be a language instruction or a target observation. 
Generally, the diffusion policy parameterized by $\phi$ is composed of an observation encoder $\E$, a diffusion transformer $\D$, and a goal encoder $\G$. The observation encoder $\E$ extracts features from the observation $\obs$, while the diffusion transformer $\D$ generates the action $\act$ conditioned on the observation $\obs$ and goal $\g$. By substituting the notations into Equation~\ref{eq:pf-ode}, diffusion policy estimates the score function $\nabla_{\act} \log p(\act | \obs, \g)$ at timestep $t$ via score matching as follows:
\begin{equation}
    \mathcal{L}_{DM} = \mathbb{E}_{\mathbf{\sigma}, \act, \boldsymbol{\epsilon}} \big[ \alpha (\sigma_t) \newline  \| \pi_{\phi}(\act_{t}, \obs, \g, \sigma_t)  - \act  \|_2^2 \big],
\end{equation}
where $\pi_{\phi} = \act + \sigma_t^2\nabla_{\act} \log p(\act | \obs, \g)$ is the neural network, $\act_{t}$ is the noised action at timestep $t$, and $\alpha (\sigma_t)$ is the loss weight. The diffusion model is trained by minimizing the score matching loss $\mathcal{L}_{DM}$, which encourages the model to generate actions that are consistent with the expert's demonstrations. In this work, we focus on accelerating the pretrained policy models by pruning and distillation algorithms, and then deploy the models on the mobile devices for real-time robot manipulation.

\subsection{Latency Analysis of Diffusion Policies}
\label{sec:method:analysis}

% Use figure* for multi-column figure
\begin{figure}[tp]
  \centering
  \includegraphics[width=\linewidth]{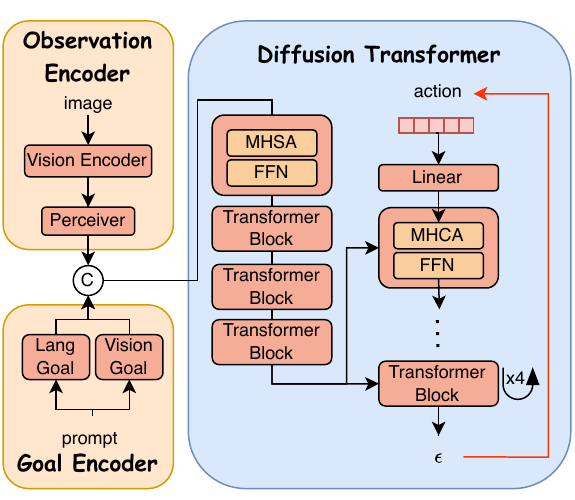}
  \vspace{-1em}
  \caption{The network architecture of MDT-V model. The model consists of three main components: the observation encoder $\E$, the goal encoder $\G$, and the diffusion transformer $\D$.}
  \label{fig:network architecture}
\end{figure}

\begin{table}[t]
  \centering
  \begin{minipage}{0.43\linewidth}
    \centering
    \resizebox{\textwidth}{!}{
    \begin{tabular}{lcc}
      \toprule
      Components & IE & DT \\
      \midrule
      Latency (ms) & 1.28 & 0.906 \\
      Parameter (M) & 11.2 & 8.97 \\
      NFE & 1 & $\underline{\mathbf{100}}$ \\
      Total Latency (ms) & 1.28 & $\underline{\mathbf{90.6}}$ \\
      \midrule
      Latency (ms) & 1.28 & 0.68 \\
      Parameter (M) & 11.2 & 4.76 \\
      NFE & 1 & $\underline{\mathbf{4}}$ \\
      Total Latency (ms) & 1.28 & $\underline{\mathbf{2.72}}$ \\ % 1.912}}$ \\
      \bottomrule
    \end{tabular}
    }
    \subcaption{DP-T Model}
  \end{minipage}%
  \begin{minipage}{0.57\linewidth}
    \centering
    \resizebox{\textwidth}{!}{
    \begin{tabular}{lccc}
      \toprule
      Components & GLE & IE & DT \\
      \midrule
      Latency (ms) & 3.74 & 3.78 & 2.25 \\
      Parameter (M) & 151.28 & 111.05 & 22.52 \\
      NFE & 1 & 2 & $\underline{\mathbf{10}}$ \\
      Total Latency (ms) & 3.74 & 7.56 & $\underline{\mathbf{22.25}}$ \\
      \midrule
      Latency (ms) & 3.74 & 3.78 & 1.025 \\
      Parameter (M) & 151.28 & 111.05 & 12.47 \\
      NFE & 1 & 2 & $\underline{\mathbf{4}}$ \\
      Total Latency (ms) & 3.74 & 7.56 & $\underline{\mathbf{4.1}}$ \\
      \bottomrule
    \end{tabular}
    }
    \subcaption{MDT-V Model}
  \end{minipage}
  % \vspace{-1em}
  \caption{Time analysis for the (a) DiffusionPolicy Transformer (DP-T) and (b) MDT-V models on iPhone 13 (\textbf{the top four rows show the original models, and the bottom four rows show the pruned models}). The device features a 16-core Apple Neural Engine capable of 16 trillion operations per second. With the aid of \emph{LightDP}, the diffusion transformers in DP-T and MDT-V achieve latency reductions from 90.6 ms and 22.25 ms to 2.72 ms and 4.1 ms, respectively. \emph{IE: Image Encoder, DT: Diffusion Transformer, GLE: Goal Language Encoder, NFE is short for the number of score function evaluations, \ie, inference steps., M: Million, ms: milliseconds}.}
  \label{tab:time analysis}
  \vspace{-1.5em}
\end{table}

Since the diffusion policy is designed for real-time robot manipulation, it is crucial to assess the on-device latency of the policy models. Given the structural similarities among these models, we use the MDT-V model as an example. As shown in Figure~\ref{fig:network architecture}, the MDT-V model supports multiple modalities of input, including an observation encoder for extracting the image features (\ie, the Voltron Network~\cite{karamcheti23voltron} for MDT-V model), a goal encoder for processing the high-level instruction (\ie, the CLIP Text Encoder), and a diffusion transformer for generating the robot action.

As shown in Table~\ref{tab:time analysis}, we evaluate the latency of the DP-T and MDT-V models on iPhone13. 
For DP-T, the network consists of two major components, the image encoder employs a ResNet18 model for converting the input image into embedding as the condition for the diffusion transformer, which costs a tiny portion of the total latency (1.28ms). The diffusion transformer is an 8-layer transformer, which is the main bottleneck of the model (90.6 ms), demands 100 iterative denoising steps to get the final action prediction. The similar observation can be found in the MDT-V model, where the Voltron network costs relatively less time (7.56ms) compared to the diffusion transformer (22.25ms), which slows down the on-device generation process. 
By breaking down the architecture of the policy models, we identify the bottleneck of the model, which is the diffusion transformer in both models. The architecture of the diffusion transformer can be formulated as a stack of $N$ transformer blocks, where each block contains a multi-head attention layer (MHA) and a feed-forward network (FFN) layer, formulated as $\layer_i = \text{FFN}(\text{MHA}(\cdot))$. Since the diffusion transformer requires multiple denoising steps to generate the action prediction, which leads to a high latency of the model. To address this issue, we propose to accelerate the model by pruning and distillation, as described in the following sections.

\subsection{Prune the Model by Learning}
\label{sec:method:pruning}

% Use figure* for multi-column figure
\begin{figure*}[tp]
  \centering
  \includegraphics[width=\linewidth]{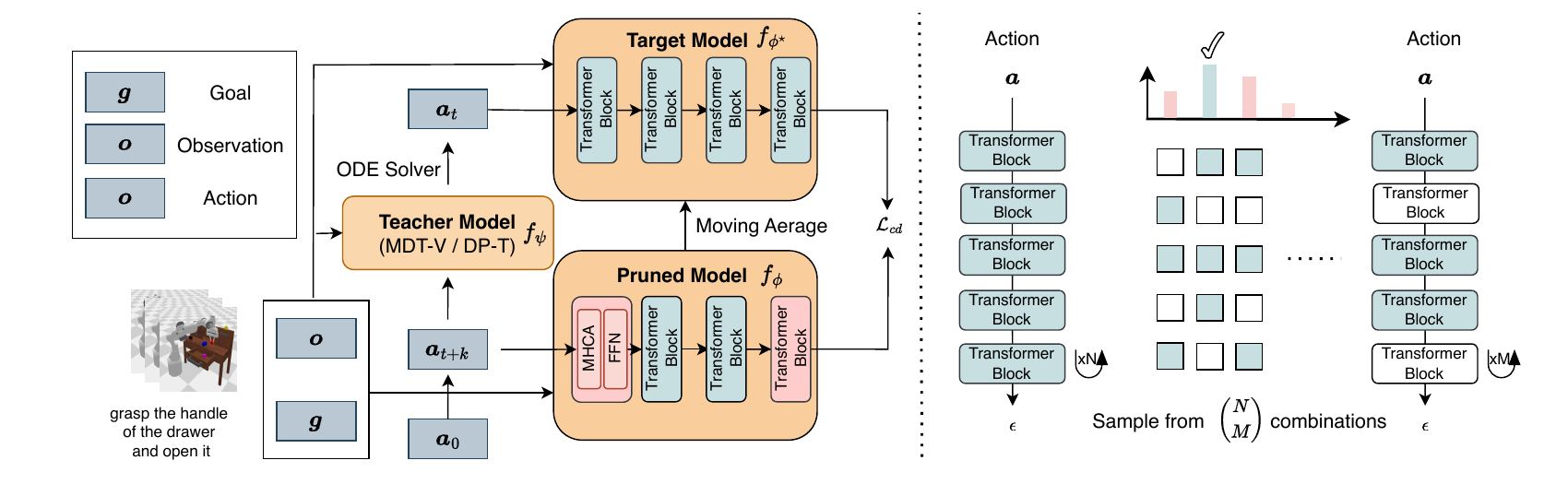}
  \caption{The training pipeline of our proposed \emph{LightDP}. In the \textbf{left} figure, we present the consistency distillation pipeline adopted in our method. The Student Model $f_{\phi}$ is initialized with the Teacher Model $f_{\psi}$ and then pruned by the learnable pruning technique introduced in Section~\ref{sec:method:pruning}. Given the sampled demonstration data $(\obs, \act, \g)$, we first add noise to obtain the noised action $\act_t$ at the timestep $t$, the Teacher Model $f_{\psi}$ is used to predict the noised action $\act_{t+k}$ at the timestep $t+k$. Then, two noised actions $\act_t$ and $\act_{t+k}$ are fed into the Student Model $f_{\phi}$ and the Target Model $f_{\phi^{\star}}$ to calculate the consistency loss. The Target Model is updated by the Student Model with a momentum update.
  In the \textbf{right} figure, we present the prune by learning technique used in our method, where a set of Bernoulli variables (gate score) is learned to perform the differentiable sampling of the pruned model, which is jointly optimized with the model parameters during the pruning process.}
  \label{fig:training pipeline}
\end{figure*}

To obtain a smaller model, we adopt the layer pruning technique to remove the redundant layers in the diffusion transformer. Given the $N$-layer diffusion transformer, we aim to find a binary mask $\mask(N) = \{m_1, m_2, ..., m_N\}$ identifying the layers to be pruned, where $m_i \in \{0, 1\}$ indicates whether this layer is retained or pruned. Conventionally, the pruning process is formulated as an optimization problem to minimize the loss $\mathcal{L}$ after pruning, which can be formulated as $\min_{\mask, {\pi_{\hat{\phi}} }} \mathbb{E}_x \left[ \mathcal{L}(x, \pi_{\phi}, \mask) \right]$, where $\pi_{\phi} =\Pi_{i=1}^N {\layer_i}$ is the vanilla model, and ${\pi}_{\hat{\phi}}$ is the model after pruning.

However, this pruning problem is NP-hard~\cite{blumensath2008iterative,frantar2023sparsegpt} since both the mask $\mask$ and weight $\hat{\phi}$ are jointly optimized. To address this, a common approach is a two-stage pruning process: first determine the mask $M$ (by minimizing the loss $L$ with a given criterion), then fine-tune the pruned model to recover performance. However, this two-step approach can be suboptimal, since the model may not fully recover performance after pruning. To address this issue, we propose to use a single-stage pruning method~\cite{fang2025tinyfusion}, where the mask $\mask$ and weight $\hat{\phi}$ are jointly optimized to minimize the loss $\mathcal{L}$ after pruning. 

Specifically, the $\mask$ is modeled as a probability distribution $\mask_i \sim \text{Bernoulli}((p_i))$, where $p_i$ is the gate score optimized during the training process. We leverage Singular Value Decomposition (SVD) to estimate layer importance, since SVD is a common technique in model compression [17,25]. Compared to alternatives like Canonical Polyadic or Kronecker product decompositions, SVD provides singular values that capture the most significant components of a weight matrix. We initialize the gate score with the SVD decomposition, which is formulated as:
\begin{equation}
    \mathcal{I}(\W) = ||\W - SVD(\W, k)|| 
                   = ||\W - \boldsymbol{U}_k \boldsymbol{S}_k \boldsymbol{V}_k^T||_F,
    \label{eqn:svd}
\end{equation}
where $\W$ is the weight matrix of the transformer block, and $SVD(\W, k)$ is the reconstructed weight matrix using the top-$k$ singular values.
Specifically, the SVD decomposition is applied to the weight matrix of each transformer block, including the query, key, and value weight matrix of the attention layer and MLP layers in the FFN module. Then, the gate score is initialized with the importance score by $p_i = \frac{\mathcal{I}{(\phi_i)}}{\sum_{i=1}^L \mathcal{I}{(\phi_i)}}$, where $\phi_i$ is the weight matrix in the $i$-th block of diffusion transformer.

As shown in Figure~\ref{fig:training pipeline}, the model is trained with a learnable gate selection mechanism via Gumbel-Softmax trick~\cite{jang2016categorical,fang2025maskllm}, which could be used to select the block to be pruned. If the $i$-th block is dropped during training, we make its output identical to its input (an identity mapping)., which could be formulated as:
\begin{equation}
    x_{i+1} = m_i\layer_i(x_i) + (1-m_i)x_i,
    \label{eqn:forward}
\end{equation}
where $x_i$ and $\layer_i(x_i)$ represent the input and output of layer $\layer_i$, respectively. The gate score is updated during the training process, which could be used to select the block to be pruned. At the end of training, to obtain an $N$ layer diffusion transformer, we select the $N$ layers with the highest gate score. To further recover the performance after pruning, we continue to fine-tune the model without adopting the mask selection process.

\subsection{Step Distillation}
\label{sec:method:distillation}
With the pruned model, the one-step inference speed could be significantly improved. However, the model still requires multiple denoising steps to obtain a high-quality action prediction, which raises a non-negligible computation cost. To address this issue, we employ the consistency distillation to train the model as a consistency model, which could achieve comparable performance with the original model but with fewer denoising steps.

As introduced in Section~\ref{sec:prelimary}, consistency distillation aims to train the model $\pi_\phi$ to satisfy the consistency property across the different noise levels, denoted as $\pi_{\phi}(\act_t, \obs, \g, \sigma_t) =  \pi_{\phi}(\act_{t^\prime}, \obs, \g, \sigma_{t^\prime})$.
The distilled model is reparameterized as EDM, which is formulated as:
\begin{equation}
    \pi_{\phi}(\act_t, \obs, \g, \sigma_t) = c_{skip}(t)\act_t+c_{out}(t)f_{\phi}(c_{in}(t)\act_t, c_{noise}(t)),
\end{equation}
where $c_{\text{skip}}$, $c_{\text{in}}$, $c_{\text{out}}$, and $c_{\text{noise}}$ satisfy the boundary condition, and $f_{\phi}$ is the distilled model.

As shown in Figure~\ref{fig:training pipeline}, the Student Model $f_{\phi}$ is initialized with the Teacher Model $f_{\psi}$ and then pruned by the learnable pruning technique introduced in Section~\ref{sec:method:pruning}. Given the sampled demonstration data $(\obs, \act, \g)$, we first add noise to obtain the noised action $\act_{t+k}$ at the timestep $t+k$, the Teacher Model $f_{\psi}$ is used to predict the noised action $\act_{t}$ at the timestep $t$. 
Then, two noised actions $\act_{t+k}$ and $\act_{t}$ are fed into the Student Model $f_{\phi}$ and the Target Model $f_{\phi^{\star}}$ to calculate the consistency loss $\mathcal{L}_{\text{CD}}$ as follows:
\begin{equation}
    \mathcal{L}_{\text{CD}} = \mathbb{E} \left[ \left\| f_{\phi}(\act_{t+k}, \obs, \g) - f_{\phi^{\star}}(\act_{t}, \obs, \g) \right\|_2^2 \right],
\end{equation}
where $\| \cdot \|_2$ is the $\ell_2$ norm. The Target Model $f_{\phi^{\star}}$ is updated with the exponential moving average (EMA) of the parameter $f_{\phi}$ defined as $f_{\phi^{\star}} \leftarrow \texttt{sg}(\mu f_{\phi^{\star}} + (1 - \mu) f_{\phi})$, where $\texttt{sg}(\cdot)$ denotes the stopgrad operation and $\mu$ satisfies $0 \leq \mu < 1$. Both Student Model and Target Model are initialized with the Teacher Model.

\section{Experiments}
\label{sec:experiments}

In this section, we introduce the experimental settings, including the baselines, benchmarks, and evaluation metrics in~Section~\ref{sec:exp:benchmarks}. And introduce the details about baselines used in our experiments, as well as the implementation details in Section~\ref{sec:exp:implementation details}. Subsequently, we present the main results and the analysis of our experiments in Section~\ref{sec:exp:dpt}.

\subsection{Benchmarks and Evaluation Metrics}
\label{sec:exp:benchmarks}
We evaluate our method on the following benchmarks:
\begin{itemize}
  \item \textbf{Push-T} was first introduced in IBC~\cite{florence2022implicit} used to evaluate the performance of Diffusion Policies. This task is designed to test the embodied agent's ability to manipulate objects with a fixed end-effector. In the task, the agent is required to push a T-shaped block into a target goal zone, which is marked by green lines in a table. The task is varied by changing the initial position of the block and the end-effector. And the task provides two types of observations: RGB images and keypoint-based states. In the experiments, we use both types of observations to evaluate the performance of our method. And we follow the evaluation protocol adopted in Diffusion Policy~\cite{chi2023diffusion} to evaluate the success rate of the manipulation task.
  \item \textbf{CALVIN~\cite{mees2022calvin}} is a simulation benchmark for measuring the performance of long-horizon language-conditioned tasks. The benchmark dataset is split into four manipulation environments, A, B, C, and D. The environments share a similar structure, like a table with objects on it, but the objects and the goal are not always the same. The agent is requested to follow the instructions to manipulate the objects on the table to achieve the goal. There are 6-hour human-teleoperated recording data in each environment, and only 1\% of the data is annotated with language instructions. We use the Average Rollout Length as the main evaluation metric in the experiments.
  \item \textbf{LIBERO~\cite{liu2023libero}} was developed for long-life robotic decision making to build the generalist agent that can perform a wide range of tasks. The benchmark comprises 130 tasks across 4 suites: LIBERO-Spatial, LIBERO-Object, LIBERO-Goal, LIBERO-100. The first three suites are designed to test the agent's ability to disentangle the transfer of declarative and procedural knowledge, while LIBERO-100 is a suite of 100 tasks with entangled knowledge transfer. 
\end{itemize}

\subsection{Implementation Details}
\label{sec:exp:implementation details}
\noindent\textbf{Base Models}. Through this work, we have mentioned both DiffusionPolicy Transformer and MDT-V in terms of their wide use in imitation learning, especially in the object manipulation tasks. As our purpose is to compress the model to make it more efficient and faster on mobile devices. We choose these two models as our base models. 
DiffusionPolicy Transformer is a transformer-based policy network that only supports image input. The model consists of a diffusion transformer and a visual encoder.

MDT is a multi-modal policy network that integrates the pre-trained multi-modal feature extractor named Voltron. We also implement MoDE, which is an MoE-based policy network that achieves the state-of-the-art performance on the CALVIN and LIBERO benchmarks.
In the experiments, we consider compressing the widely used Diffusion Policies, including Diffusion-Policy-T~\cite{chi2023diffusion}, and MDT~\cite{reuss24mdt}.
Diffusion-Policy-T~\cite{chi2023diffusion} is a transformer-based policy network for imitation learning that supports only image input. 
MDT~\cite{reuss24mdt}, by integrating the pre-trained multi-modal feature extractor named Voltron~\cite{karamcheti23voltron}, MDT has achieved good results on the CALVIN dataset. 

\noindent\textbf{Implementation Details}.\label{sec:exp:implement} Our implementation is based on PyTorch. We conducted training on NVIDIA RTX 3090 and H800 GPUs. Then, we converted the model trained on GPU to Core ML model format (mlpackage, based on Apple's ml-stable-diffusion) and measured latency in Xcode Instruments on an iPhone 13 (A15 Bionic, iOS 18.3.1). For network pruning, we adopt the local block pruning scheme from TinyFusion~\cite{fang2025tinyfusion} to build up a local block with scheme $N$:$M$. In this $N$:$M$ scheme, each group of $M$ consecutive layers (a `block') is pruned down to $N$ layers.. For instance, when we keep $N=3$ layers from a local block with $M=4$ layers in total, we have $\binom{4}{3}=4$ choices, corresponding to $\mask = [[1,1,1,0],[1,1,0,1],[1,0,1,1],[0,1,1,1]]$. 
Our consistency distillation is applied to the model's $x_0$ prediction (predicting the denoised action), following common practice, and we start the EMA decay rate at 0.95 and gradually increase it to 0.999 over the course of training to stabilize the Target model updates. We use the DDIM Solver~\cite{song2020denoising} for distillation, with a skip interval of 10 steps (i.e., distill every 10th diffusion step). 
We keep the most hyper-parameters consistent with the original implementation of the base models. For DP-T, the input is a hybrid of RGB image and low-dimension state, the size of image is $84\times 84$, and the observation sequence length is set as 2, the transformer block of the diffusion transformer is with the hidden size of 256, the number of heads is 4, and the number of layers in DP-T is 8. For MDT, the input is multi-modal, which includes two RGB images at different views as observation and a language instruction as the goal. We adopt AdamW as the optimizer with a learning rate of $1e-4$, and the batch size is set as 64. We train the model for 30 epochs on the CALVIN datasets, within the last epochs, the Student Model $f_{\phi}$ is pruned based on the gate score at $20$-th epoch.

\subsection{Evaluation on DiffusionPolicy Transformer}
\label{sec:exp:dpt}
In this section, we conduct the experiments based on DP-T as reported in Table~\ref{tab:exp_dp}, we can find that the pruned model can achieve a comparable success rate with the original model, but with a smaller model size and faster inference speed.

\begin{table*}[htbp]
  \centering
  \resizebox{0.85\textwidth}{!}{
  \begin{tabular}{lcccccc}
  \toprule
  Method & Depth & Param (M) & NFE & GFLOPs & Inference Speed (ms) & Success Rate \\ \midrule
  DP-T & \multirow{2}{*}{8} & \multirow{2}{*}{8.97} & \multirow{2}{*}{100} & \multirow{2}{*}{4.39} & \multirow{2}{*}{90.6} & 0.772±0.039 \\
  $\text{DP-T}^{\star}$ & & & & & & 0.754±0.023 \\ \midrule
  DP-T-D6/6-8 & \multirow{2}{*}{6} & \multirow{2}{*}{6.87} & \multirow{2}{*}{4} & \multirow{2}{*}{0.134} & \multirow{2}{*}{4.79} & 0.752±0.019 \\
  DP-T-D6/4-4 & & & & & & 0.732±0.034 \\ \midrule
  DP-T-D4/4-8 & \multirow{3}{*}{4} & \multirow{3}{*}{4.76} & \multirow{3}{*}{4} & \multirow{3}{*}{0.091} & \multirow{3}{*}{2.72} & 0.747±0.010 \\
  DP-T-D4/2-4 & & & & & & 0.732±0.013 \\
  DP-T-D4/1-2 & & & & & & 0.757±0.018 \\ \midrule
  DP-T-D2/2-8 & \multirow{2}{*}{2} & \multirow{2}{*}{2.65} & \multirow{2}{*}{4} & \multirow{2}{*}{0.049} & \multirow{2}{*}{0.97} & 0.730±0.022 \\
  DP-T-D2/1-4 & & & & & & 0.724±0.030 \\
  \bottomrule
  \end{tabular}
  }
\caption{Performance comparison of \emph{LightDP} compressed models with varying depth and inference steps. All models are trained on the same Push-T dataset for 3K epochs. $\text{DP-T}^{\star}$ refers to the baseline model evaluated by us. DP-T-D$\textcolor{red}{L}$/$\textcolor{red}{N}$-$\textcolor{red}{M}$ indicates that $\textcolor{red}{L}$ blocks are retained during the pruning process, with a local block scheme of $N$:$M$. NFE is short for the number of score function evaluations, \ie, inference steps. Detailed experiments on the Robomimic dataset are provided in Section~\ref{sec:supp:extensive_exp_dpt}.}
\label{tab:exp_dp}
\end{table*}

\noindent\textbf{Quantitative Results}. The vanilla DP-T model contains 8 transformer blocks with alternative Multi-head Cross-Attention layers and Feed-Forward layers. The model is first trained to obtain an optimal pruning mask with the network weight updated jointly, then the model is pruned and trained via a consistency distillation loss. In our setting, we compress the model into 2, 4, and 6 layers. The results show that through our method, the pruned model can achieve a comparable success rate with the vanilla model. As we discuss in When $N$:$M$=1:2, each two successive blocks are grouped with one block pruned. With the same depth, we observe that when the capacity $M$ of the block is reduced, the performance will be slightly reduced, since large $M$ can provide more diverse pruning choices. Besides, from the perspective of the depth of the pruned model, we find that the performance of the larger depth model remains better than the smaller one, which is consistent with the intuition, but the performance gap is not significant. Especially, we find a 2-layer diffusion transformer can achieve a success rate with 0.724, which is quite close to the original model with 0.754. 
In contrast, the latency of the pruned model is greatly diminished when compared to the DP-T model. With the number of inference steps cut down to 4 and the depth limited to 2, we attain approximately 93 times speed improvement, and the FLOPs are decreased by 89.6\%. These results indicate that our proposed \emph{LightDP} successfully compresses the model while preserving the original model's performance.

\begin{table*}[htbp]
  \centering
  \resizebox{\textwidth}{!}{
  \begin{tabular}{lcccccccccc}
  \hline
  \multirow{2}{*}{Training $\rightarrow$ Test} & \multirow{2}{*}{Method} & \multirow{2}{*}{Param (M)} & \multirow{2}{*}{GFLOPs} & \multirow{2}{*}{Latency (ms)} & \multicolumn{6}{c}{Instructions in a Row (1000 chains)} \\ \cline{6-11}
   & & & & & 1 & 2 & 3 & 4 & 5 & Average Length \\ \hline
  \multirow{4}{*}{ABCD$\rightarrow$D} & MDT-V & 22.52 & 1.21 & 22.25 & 98.6\% & 95.8\% & 91.6\% & 86.2\% & 80.1\% & 4.52±(0.02) \\
   & MDT-V/E3-D3 & 17.50 & 0.36 & 8.7 & 98.3\% & 94.6\% & 91.5\% & 85.8\% & 79.6\% & 4.50±(0.06) \\
   & MDT-V/E2-D2 & 12.47 & 0.25 & 4.1 & 95.1\% & 87.9\% & 80.5\% & 71.9\% & 64.1\% & 3.94±(0.08) \\
   & MDT-V/E1-D1 & 7.45 & 0.13 & 3.39 & 92.3\% & 85.4\% & 77.2\% & 65.9\% & 61.4\% & 3.44±(0.05) \\ \hline
  \multirow{4}{*}{D$\rightarrow$D} & MDT-V & 22.52 & 1.21 & 22.25 & 93.7\% & 84.5\% & 74.1\% & 64.4\% & 55.6\% & 3.72±(0.06) \\
   & MDT-V/E3-D3 & 17.50 & 0.36 & 8.7 & 92.4\% & 82.1\% & 71.2\% & 60.5\% & 52.2\% & 3.65±(0.05) \\
   & MDT-V/E2-D2 & 12.47 & 0.25 & 4.1 & 87.1\% & 71.2\% & 58.7\% &  48.3\% & 37.9\%   & 3.00±(0.03)  \\
   & MDT-V/E1-D1 & 7.45 & 0.13 & 3.39 & 79.9\% & 63.2\% & 47.8\% & 35.0\% & 23.1\% & 2.48±(0.07) \\ \hline
\end{tabular}}
\caption{Performance comparison of \emph{LightDP} compressed MDT-V models with different depth and inference steps. All models are trained on the CALVIN D or CALVIN ABCD for 30 epochs, and then tested on the CALVIN D dataset.}
\label{tab:exp_mdt_calvin}
\vspace{-1.5em}
\end{table*}

\subsection{Evaluation on MDT-V}
\label{sec:exp:mdt}
\begin{table}[htbp]
  \centering
  \resizebox{0.47\textwidth}{!}{
  \begin{tabular}{lcccccc}
  \toprule
  Task & Spatial & Object & Goal & Long & 90 & Average \\ \midrule
  MDT-V & 78.5$\pm$1.5 & 87.5$\pm$0.9 & 73.5$\pm$2.0 & 64.8$\pm$1.5 & 67.2$\pm$1.1 & 74.3$\pm$9.1 \\
  MDT-V/E3-D3 & 77.9$\pm$1.9 & 86.5$\pm$2.1 & 71.5$\pm$3.1 & 63.2$\pm$2.3 & 66.8$\pm$0.8 & 73.2$\pm$9.9 \\ \bottomrule
  \end{tabular}}
  \caption{Performance comparison of \emph{LightDP} compressed MDT-V/E3-D3 model on the benchmark LIBERO. For each task, the achieved score is presented along with its variability (mean$\pm$standard deviation)}
  
\label{tab:exp_evaluation_mdt_libero}
\end{table}
In this section, we conduct the experiments based on MDT-V as reported in Table~\ref{tab:exp_mdt_calvin}. Since MDT-V consists of 4-layer TransformerEncoder and 4-layer TransformerDecoder, we keep the number of encoder layers the same as the decoder layers, therefore, we compress the model into 2, 4, and 6 layers as well as DP-T. Compared with the original model, the 6-layer model achieves comparable performance, while the 4-layer model has a significant performance drop and the 2-layer model has the worst performance. The results show the MDT-V model is more compact than the DP-T model. 
In addition, as detailed in Table~\ref{tab:exp_mdt_calvin}, the ABCD$\rightarrow$D results reveal that the full MDT model attains very high success percentages across the chain (e.g., 98.6\% on the first instruction, gradually decreasing to 80.1\%), with an average chain length of 4.52. In contrast, the pruned variants show a noticeable decline in performance, where MDT-V/E1-D1, for instance, achieves only 92.3\% initially and drops to 61.4\%, with a reduced average chain length of 3.44. Similarly, in the D→D scenario, all models register lower performance, with the most compressed model suffering from a steep decline in both success rate and average chain length. These observations underscore the trade-off between model compactness and performance, highlighting that even a slight reduction in network depth can substantially impact the ability to sustain performance over extended inference sequences.
Besides, we also conduct the experiments on the LIBERO datasets shown in Table~\ref{tab:exp_evaluation_mdt_libero}, by comparing MDT-V and MDT-V/E3-D3 across LIBERO task suites, we find that the pruned model achieves comparable performance with the original model. On average, while MDT-V shows a marginally better overall result with less variability, the inference speed and model size are reduced significantly, which could be beneficial for deployment on mobile devices.

\subsection{Ablation Study}
\label{sec:exp:ablation}

\begin{table}[htbp]
  \centering
  \vspace{-1em}
  \resizebox{0.47\textwidth}{!}{
      \begin{tabular}{lccccc}
      \hline
      {Method} & {Param (M)} & {GFLOPs} & {Latency (ms)} & {Average Length} \\\hline
      MDT-V & 22.52 & 1.21 & 22.25 & 3.72$\pm$(0.06) \\
      MDT-V w/ prune & 17.50 & 0.91 & 18.87 & 3.70$\pm$(0.08) \\
      MDT-V w/ CD & 22.52 & 0.48 & 11.34 & 3.69$\pm$(0.02) \\
      MDT-V/E3-D3 & 17.50 & 0.36 & 8.70 & 3.65$\pm$(0.05) \\ \hline
      \end{tabular}
  }
    \caption{Ablation study on the effect of the proposed learnable pruning and step distillation based on MDT-V, the performance is evaluated on the CALVIN D$\rightarrow$D task suite. w/ prune means learnable pruning technique, and w/ CD means step distillation. MDT-V/E3-D3 combines learnable pruning and step distillation.}
  \label{tab:exp_ablation_mdt_calvin}
\end{table}

In this section, we ablate the effectiveness of the proposed method by removing the consistency distillation and learnable pruning. As shown in Table~\ref{tab:exp_ablation_mdt_calvin}, when learnable pruning is applied (MDT-V w/prune), we observe a reduction in the number of parameters and GFLOPs, along with slightly reduced latency (from 22.25ms to 18.87ms), while preserving similar behavior in the generated actions. Likewise, employing consistency distillation (MDT-V w/CD) considerably reduces the GFLOPs and latency with only minimal reduction in the average rollout length. Notably, the combined approach (MDT-V/E3-D3) delivers the best trade-off by minimizing latency and computational cost, thereby demonstrating the efficiency of our design modifications without significant degradation in performance.

\subsection{Qualitative Results}
\label{sec:exp:qualitative}
% Use figure* for multi-column figure
\begin{figure}[tp]
  \centering
  \includegraphics[width=\linewidth]{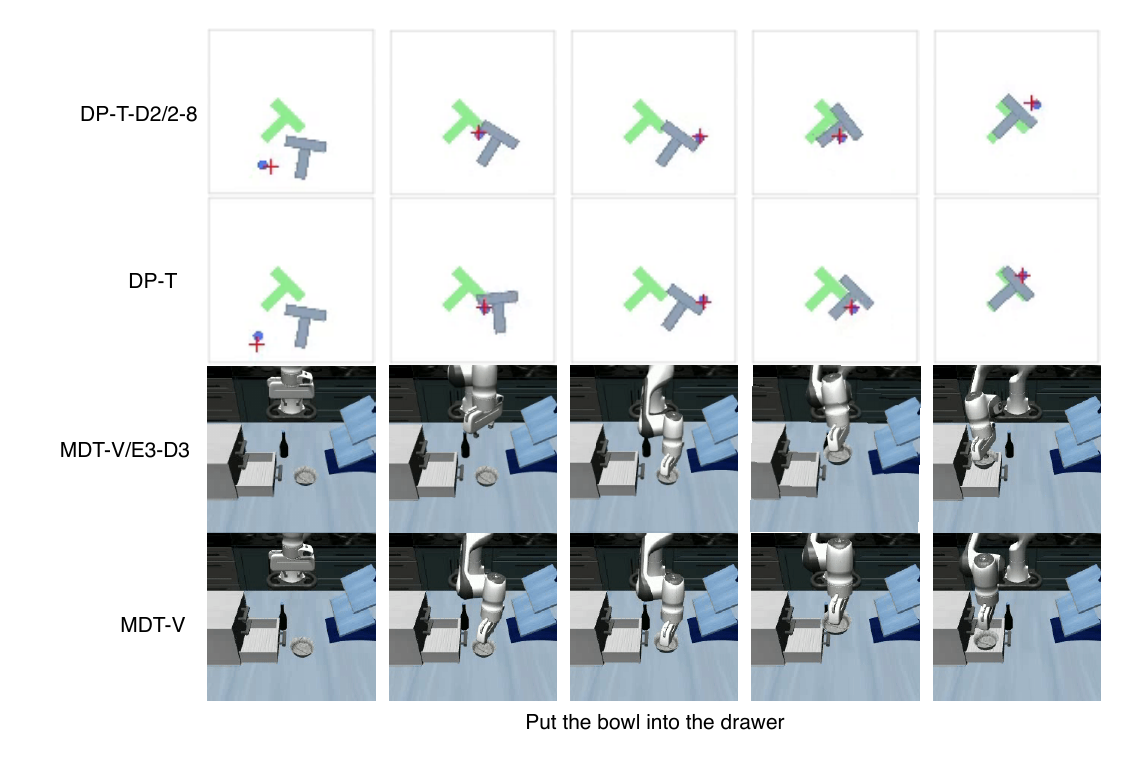}
  \caption{Qualitative comparison of the pruned models and original models. We observe that the pruned models can mimic the behaviors of the original models, which demonstrates the step distillation process is capable of transferring the knowledge from the original model to the pruned model.}
  \label{fig:exp_vis}
  \vspace{-1.5em}
\end{figure}

Figure~\ref{fig:exp_vis} displays rollout of the pruned DP-T and MDT-V models on the Push-T and LIBERO tasks. In the Push-T task, the pruned model successfully pushed the T-shaped block into the goal zone, without any failure in the manipulation process. And in the LIBERO task suite that requires the agent to follow the instructions to manipulate the objects on the table to achieve the goal, the pruned model can also successfully complete the task. 
By adopting \emph{LightDP} on the original DP-T and MDT-V models, we obtain the lightweight policy models. Here we present the visual comparison between the pruned model and the original model. With the rollouts in the Push-T task and the CALVIN tasks. In Figure~\ref{fig:exp_vis}, the upper two rows present the pruned model DP-T-D2/2-8 and DP-T on the Push-T task, and the bottom two rows show the pruned model MDT-V/E3-D3 and the original model MDT-V. We observe that the pruned models can mimic the behaviors of the original models, which demonstrates the step distillation process is capable of transferring the knowledge from the original model to the pruned model.
Except for the experiments on simulation environments, we also conduct the real-world experiments on robotic arms as presented in Section~\ref{sec:supp:real-world_exp}. The results show that the pruned model can achieve a comparable success rate with the original model, which demonstrates the effectiveness of our method in real-world scenarios.

\section{Conclusion and Limitation}
\label{sec:conclusion}
In this paper, we introduced the \textit{LightDP} framework, aiming at accelerating Diffusion Policies on the mobile devices. Specifically, we analyze the architecture of the widely-used DP-T and MDT-V baselines, observe the iterative denoising process, and the high cost of the network inference hurdles the real-time application of these models on the mobile robots. To address this issue, we employed two strategies: 1) adopting a lightweight network architecture via a learnable pruning method, and 2) reducing the number of inference steps to speed up the denoising process. We have benchmarked the proposed \emph{LightDP} framework on Push-T, Robomimic, CALVIN, and LIBERO datasets, demonstrating a significant improvement in terms of inference speed and memory consumption.

\noindent\textbf{Limitations}. In this paper, we mainly focus on the Diffusion Policies, while the new proposed VLA models are not well explored in this work. We leave this as the future work.

{
    \small
    \bibliographystyle{ieeenat_fullname}
    \bibliography{main}

\begin{thebibliography}{53}
\providecommand{\natexlab}[1]{#1}
\providecommand{\url}[1]{\texttt{#1}}
\expandafter\ifx\csname urlstyle\endcsname\relax
  \providecommand{\doi}[1]{doi: #1}\else
  \providecommand{\doi}{doi: \begingroup \urlstyle{rm}\Url}\fi

\bibitem[Black et~al.(2024)Black, Brown, Driess, Esmail, Equi, Finn, Fusai, Groom, Hausman, Ichter, Jakubczak, Jones, Ke, Levine, Li-Bell, Mothukuri, Nair, Pertsch, Shi, Tanner, Vuong, Walling, Wang, and Zhilinsky]{black2024pi0}
Kevin Black, Noah Brown, Danny Driess, Adnan Esmail, Michael Equi, Chelsea Finn, Niccolo Fusai, Lachy Groom, Karol Hausman, Brian Ichter, Szymon Jakubczak, Tim Jones, Liyiming Ke, Sergey Levine, Adrian Li-Bell, Mohith Mothukuri, Suraj Nair, Karl Pertsch, Lucy~Xiaoyang Shi, James Tanner, Quan Vuong, Anna Walling, Haohuan Wang, and Ury Zhilinsky.
\newblock $\pi_0$: A vision-language-action flow model for general robot control, 2024.

\bibitem[Black et~al.(2025)Black, Galliker, and Levine]{black2025real}
Kevin Black, Manuel~Y Galliker, and Sergey Levine.
\newblock Real-time execution of action chunking flow policies.
\newblock \emph{arXiv preprint arXiv:2506.07339}, 2025.

\bibitem[Blumensath and Davies(2008)]{blumensath2008iterative}
Thomas Blumensath and Mike~E Davies.
\newblock Iterative thresholding for sparse approximations.
\newblock \emph{Journal of Fourier analysis and Applications}, 14:\penalty0 629--654, 2008.

\bibitem[Brohan et~al.(2023)Brohan, Brown, Carbajal, Chebotar, Dabis, Finn, Gopalakrishnan, Hausman, Herzog, Hsu, Ibarz, Ichter, Irpan, Jackson, Jesmonth, Joshi, Julian, Kalashnikov, Kuang, Leal, Lee, Levine, Lu, Malla, Manjunath, Mordatch, Nachum, Parada, Peralta, Perez, Pertsch, Quiambao, Rao, Ryoo, Salazar, Sanketi, Sayed, Singh, Sontakke, Stone, Tan, Tran, Vanhoucke, Vega, Vuong, Xia, Xiao, Xu, Xu, Yu, and Zitkovich]{brohan23rt1}
Anthony Brohan, Noah Brown, Justice Carbajal, Yevgen Chebotar, Joseph Dabis, Chelsea Finn, Keerthana Gopalakrishnan, Karol Hausman, Alexander Herzog, Jasmine Hsu, Julian Ibarz, Brian Ichter, Alex Irpan, Tomas Jackson, Sally Jesmonth, Nikhil~J. Joshi, Ryan Julian, Dmitry Kalashnikov, Yuheng Kuang, Isabel Leal, Kuang{-}Huei Lee, Sergey Levine, Yao Lu, Utsav Malla, Deeksha Manjunath, Igor Mordatch, Ofir Nachum, Carolina Parada, Jodilyn Peralta, Emily Perez, Karl Pertsch, Jornell Quiambao, Kanishka Rao, Michael~S. Ryoo, Grecia Salazar, Pannag~R. Sanketi, Kevin Sayed, Jaspiar Singh, Sumedh Sontakke, Austin Stone, Clayton Tan, Huong~T. Tran, Vincent Vanhoucke, Steve Vega, Quan Vuong, Fei Xia, Ted Xiao, Peng Xu, Sichun Xu, Tianhe Yu, and Brianna Zitkovich.
\newblock {RT-1:} robotics transformer for real-world control at scale.
\newblock In \emph{Robotics: Science and Systems}, 2023.

\bibitem[Castells et~al.(2024)Castells, Song, Piao, Choi, Kim, Yim, Lee, Kim, and Kim]{castells2024edgefusion}
Thibault Castells, Hyoung-Kyu Song, Tairen Piao, Shinkook Choi, Bo-Kyeong Kim, Hanyoung Yim, Changgwun Lee, Jae~Gon Kim, and Tae-Ho Kim.
\newblock Edgefusion: On-device text-to-image generation.
\newblock \emph{arXiv preprint arXiv:2404.11925}, 2024.

\bibitem[Chen et~al.(2023)Chen, Relic, Azevedo, Zhang, Gross, Xu, Zhou, and Schroers]{chen2023neural}
Zhenghao Chen, Lucas Relic, Roberto Azevedo, Yang Zhang, Markus Gross, Dong Xu, Luping Zhou, and Christopher Schroers.
\newblock Neural video compression with spatio-temporal cross-covariance transformers.
\newblock In \emph{Proceedings of the 31st ACM International Conference on Multimedia}, pages 8543--8551, 2023.

\bibitem[Chen et~al.(2024)Chen, Zhou, Hu, and Xu]{chen2024group}
Zhenghao Chen, Luping Zhou, Zhihao Hu, and Dong Xu.
\newblock Group-aware parameter-efficient updating for content-adaptive neural video compression.
\newblock In \emph{Proceedings of the 32nd ACM International Conference on Multimedia}, pages 11022--11031, 2024.

\bibitem[Chi et~al.(2023)Chi, Xu, Feng, Cousineau, Du, Burchfiel, Tedrake, and Song]{chi2023diffusion}
Cheng Chi, Zhenjia Xu, Siyuan Feng, Eric Cousineau, Yilun Du, Benjamin Burchfiel, Russ Tedrake, and Shuran Song.
\newblock Diffusion policy: Visuomotor policy learning via action diffusion.
\newblock \emph{The International Journal of Robotics Research}, 2023.

\bibitem[Fang et~al.(2023)Fang, Ma, Song, Mi, and Wang]{fang2023depgraph}
Gongfan Fang, Xinyin Ma, Mingli Song, Michael~Bi Mi, and Xinchao Wang.
\newblock Depgraph: Towards any structural pruning.
\newblock In \emph{CVPR}, 2023.

\bibitem[Fang et~al.(2025{\natexlab{a}})Fang, Li, Ma, and Wang]{fang2025tinyfusion}
Gongfan Fang, Kunjun Li, Xinyin Ma, and Xinchao Wang.
\newblock Tinyfusion: Diffusion transformers learned shallow.
\newblock In \emph{CVPR}, pages 18144--18154, 2025{\natexlab{a}}.

\bibitem[Fang et~al.(2025{\natexlab{b}})Fang, Yin, Muralidharan, Heinrich, Pool, Kautz, Molchanov, and Wang]{fang2025maskllm}
Gongfan Fang, Hongxu Yin, Saurav Muralidharan, Greg Heinrich, Jeff Pool, Jan Kautz, Pavlo Molchanov, and Xinchao Wang.
\newblock Maskllm: Learnable semi-structured sparsity for large language models.
\newblock \emph{Advances in Neural Information Processing Systems}, 37:\penalty0 7736--7758, 2025{\natexlab{b}}.

\bibitem[Florence et~al.(2022)Florence, Lynch, Zeng, Ramirez, Wahid, Downs, Wong, Lee, Mordatch, and Tompson]{florence2022implicit}
Pete Florence, Corey Lynch, Andy Zeng, Oscar~A Ramirez, Ayzaan Wahid, Laura Downs, Adrian Wong, Johnny Lee, Igor Mordatch, and Jonathan Tompson.
\newblock Implicit behavioral cloning.
\newblock In \emph{Conference on robot learning}, pages 158--168. PMLR, 2022.

\bibitem[Frankle and Carbin(2019)]{frankle2018lottery}
Jonathan Frankle and Michael Carbin.
\newblock The lottery ticket hypothesis: Finding sparse, trainable neural networks.
\newblock In \emph{ICLR}, 2019.

\bibitem[Frantar and Alistarh(2023)]{frantar2023sparsegpt}
Elias Frantar and Dan Alistarh.
\newblock Sparsegpt: Massive language models can be accurately pruned in one-shot.
\newblock In \emph{International Conference on Machine Learning}, pages 10323--10337. PMLR, 2023.

\bibitem[Fu et~al.(2024)Fu, Zhao, and Finn]{fu2024mobile}
Zipeng Fu, Tony~Z. Zhao, and Chelsea Finn.
\newblock Mobile aloha: Learning bimanual mobile manipulation with low-cost whole-body teleoperation.
\newblock In \emph{Conference on Robot Learning}, 2024.

\bibitem[Hadji et~al.(2025)Hadji, Noroozi, Escorcia, Zaganidis, Martinez, and Tzimiropoulos]{hadji2025edge}
Isma Hadji, Mehdi Noroozi, Victor Escorcia, Anestis Zaganidis, Brais Martinez, and Georgios Tzimiropoulos.
\newblock Edge-sd-sr: Low latency and parameter efficient on-device super-resolution with stable diffusion via bidirectional conditioning.
\newblock In \emph{CVPR}, pages 12789--12798, 2025.

\bibitem[Han et~al.(2015)Han, Pool, Tran, and Dally]{han2015learning}
Song Han, Jeff Pool, John Tran, and William~J Dally.
\newblock Learning both weights and connections for efficient neural network.
\newblock In \emph{NeurIPS}, 2015.

\bibitem[Hinton et~al.(2014)Hinton, Vinyals, and Dean]{hinton2015distilling}
Geoffrey Hinton, Oriol Vinyals, and Jeff Dean.
\newblock Distilling the knowledge in a neural network.
\newblock In \emph{NIPS Workshop}, 2014.

\bibitem[Ho and Salimans(2021)]{ho2021classifier}
Jonathan Ho and Tim Salimans.
\newblock Classifier-free diffusion guidance.
\newblock In \emph{NeurIPS 2021 Workshop on Deep Generative Models and Downstream Applications}, 2021.

\bibitem[Jang et~al.(2016)Jang, Gu, and Poole]{jang2016categorical}
Eric Jang, Shixiang Gu, and Ben Poole.
\newblock Categorical reparameterization with gumbel-softmax.
\newblock \emph{arXiv preprint arXiv:1611.01144}, 2016.

\bibitem[Karamcheti et~al.(2023)Karamcheti, Nair, Chen, Kollar, Finn, Sadigh, and Liang]{karamcheti23voltron}
Siddharth Karamcheti, Suraj Nair, Annie~S. Chen, Thomas Kollar, Chelsea Finn, Dorsa Sadigh, and Percy Liang.
\newblock Language-driven representation learning for robotics.
\newblock In \emph{Robotics: Science and Systems}, 2023.

\bibitem[Karras et~al.(2022)Karras, Aittala, Aila, and Laine]{karras2022elucidating}
Tero Karras, Miika Aittala, Timo Aila, and Samuli Laine.
\newblock Elucidating the design space of diffusion-based generative models.
\newblock pages 26565--26577, 2022.

\bibitem[Kim et~al.(2024{\natexlab{a}})Kim, Song, Castells, and Choi]{kimBKSDMLightweightFast2023}
Bo-Kyeong Kim, Hyoung-Kyu Song, Thibault Castells, and Shinkook Choi.
\newblock Bk-sdm: A lightweight, fast, and cheap version of stable diffusion.
\newblock In \emph{ECCV}, 2024{\natexlab{a}}.

\bibitem[Kim et~al.(2024{\natexlab{b}})Kim, Pertsch, Karamcheti, Xiao, Balakrishna, Nair, Rafailov, Foster, Sanketi, Vuong, Kollar, Burchfiel, Tedrake, Sadigh, Levine, Liang, and Finn]{kim2024openvla}
Moo~Jin Kim, Karl Pertsch, Siddharth Karamcheti, Ted Xiao, Ashwin Balakrishna, Suraj Nair, Rafael Rafailov, Ethan~P Foster, Pannag~R Sanketi, Quan Vuong, Thomas Kollar, Benjamin Burchfiel, Russ Tedrake, Dorsa Sadigh, Sergey Levine, Percy Liang, and Chelsea Finn.
\newblock Open{VLA}: An open-source vision-language-action model.
\newblock In \emph{Conference on Robot Learning}, 2024{\natexlab{b}}.

\bibitem[Li et~al.(2017)Li, Kadav, Durdanovic, Samet, and Graf]{li2017pruning}
Hao Li, Asim Kadav, Igor Durdanovic, Hanan Samet, and Hans~Peter Graf.
\newblock Pruning filters for efficient convnets.
\newblock In \emph{ICLR}, 2017.

\bibitem[Li et~al.(2024)Li, Liu, Zhang, Yu, Xu, Wu, Cheang, Jing, Zhang, Liu, et~al.]{li24roboflamingo}
Xinghang Li, Minghuan Liu, Hanbo Zhang, Cunjun Yu, Jie Xu, Hongtao Wu, Chilam Cheang, Ya Jing, Weinan Zhang, Huaping Liu, et~al.
\newblock Vision-language foundation models as effective robot imitators.
\newblock In \emph{ICLR}, 2024.

\bibitem[Li et~al.(2023)Li, Wang, Jin, Hu, Chemerys, Fu, Wang, Tulyakov, and Ren]{liSnapFusionTextImageDiffusion2023}
Yanyu Li, Huan Wang, Qing Jin, Ju Hu, Pavlo Chemerys, Yun Fu, Yanzhi Wang, Sergey Tulyakov, and Jian Ren.
\newblock Snapfusion: Text-to-image diffusion model on mobile devices within two seconds.
\newblock In \emph{NeurIPS}, 2023.

\bibitem[Liu et~al.(2023)Liu, Zhu, Gao, Feng, Liu, Zhu, and Stone]{liu2023libero}
Bo Liu, Yifeng Zhu, Chongkai Gao, Yihao Feng, Qiang Liu, Yuke Zhu, and Peter Stone.
\newblock Libero: Benchmarking knowledge transfer for lifelong robot learning.
\newblock \emph{Advances in Neural Information Processing Systems}, 36:\penalty0 44776--44791, 2023.

\bibitem[Luo et~al.(2023)Luo, Tan, Huang, Li, and Zhao]{luoLATENTCONSISTENCYMODELS2023}
Simian Luo, Yiqin Tan, Longbo Huang, Jian Li, and Hang Zhao.
\newblock Latent consistency models: Synthesizing high-resolution images with few-step inference.
\newblock \emph{arXiv preprint arXiv:2310.04378}, 2023.

\bibitem[Mees et~al.(2022)Mees, Hermann, Rosete-Beas, and Burgard]{mees2022calvin}
Oier Mees, Lukas Hermann, Erick Rosete-Beas, and Wolfram Burgard.
\newblock Calvin: A benchmark for language-conditioned policy learning for long-horizon robot manipulation tasks.
\newblock \emph{IEEE Robotics and Automation Letters}, 7\penalty0 (3):\penalty0 7327--7334, 2022.

\bibitem[Men et~al.(2024)Men, Xu, Zhang, Wang, Lin, Lu, Han, and Chen]{men2024shortgpt}
Xin Men, Mingyu Xu, Qingyu Zhang, Bingning Wang, Hongyu Lin, Yaojie Lu, Xianpei Han, and Weipeng Chen.
\newblock Shortgpt: Layers in large language models are more redundant than you expect.
\newblock \emph{arXiv preprint arXiv:2403.03853}, 2024.

\bibitem[Meng et~al.(2023)Meng, Gao, Kingma, Ermon, Ho, and Salimans]{meng2023distillation}
Chenlin Meng, Ruiqi Gao, Diederik~P Kingma, Stefano Ermon, Jonathan Ho, and Tim Salimans.
\newblock On distillation of guided diffusion models.
\newblock In \emph{CVPR}, 2023.

\bibitem[Molchanov et~al.(2017)Molchanov, Tyree, Karras, Aila, and Kautz]{molchanov2017pruning}
Pavlo Molchanov, Stephen Tyree, Tero Karras, Timo Aila, and Jan Kautz.
\newblock Pruning convolutional neural networks for resource efficient inference.
\newblock In \emph{ICLR}, 2017.

\bibitem[Mozer and Smolensky(1988)]{mozer1988skeletonization}
Michael~C Mozer and Paul Smolensky.
\newblock Skeletonization: A technique for trimming the fat from a network via relevance assessment.
\newblock In \emph{NeurIPS}, 1988.

\bibitem[Reuss et~al.(2023)Reuss, Li, Jia, and Lioutikov]{reuss2023goalconditionedimitationlearningusing}
Moritz Reuss, Maximilian Li, Xiaogang Jia, and Rudolf Lioutikov.
\newblock Goal-conditioned imitation learning using score-based diffusion policies, 2023.

\bibitem[Reuss et~al.(2024)Reuss, Yagmurlu, Wenzel, and Lioutikov]{reuss24mdt}
Moritz Reuss, {\"{O}}mer~Erdin{\c{c}} Yagmurlu, Fabian Wenzel, and Rudolf Lioutikov.
\newblock Multimodal diffusion transformer: Learning versatile behavior from multimodal goals.
\newblock In \emph{Robotics: Science and Systems}, 2024.

\bibitem[Reuss et~al.(2025)Reuss, Pari, Agrawal, and Lioutikov]{reuss2025mode}
Moritz Reuss, Jyothish Pari, Pulkit Agrawal, and Rudolf Lioutikov.
\newblock Efficient diffusion transformer policies with mixture of expert denoisers for multitask learning.
\newblock In \emph{ICLR}, 2025.

\bibitem[Salimans and Ho(2022)]{salimans2022progressive}
Tim Salimans and Jonathan Ho.
\newblock Progressive distillation for fast sampling of diffusion models.
\newblock In \emph{ICLR}, 2022.

\bibitem[Shukor et~al.(2025)Shukor, Aubakirova, Capuano, Kooijmans, Palma, Zouitine, Aractingi, Pascal, Russi, Marafioti, et~al.]{shukor2025smolvla}
Mustafa Shukor, Dana Aubakirova, Francesco Capuano, Pepijn Kooijmans, Steven Palma, Adil Zouitine, Michel Aractingi, Caroline Pascal, Martino Russi, Andres Marafioti, et~al.
\newblock Smolvla: A vision-language-action model for affordable and efficient robotics.
\newblock \emph{arXiv preprint arXiv:2506.01844}, 2025.

\bibitem[Song et~al.(2020)Song, Meng, and Ermon]{song2020denoising}
Jiaming Song, Chenlin Meng, and Stefano Ermon.
\newblock Denoising diffusion implicit models.
\newblock \emph{arXiv preprint arXiv:2010.02502}, 2020.

\bibitem[Song and Dhariwal(2024)]{song2024improved}
Yang Song and Prafulla Dhariwal.
\newblock Improved techniques for training consistency models.
\newblock In \emph{ICLR}, 2024.

\bibitem[Song et~al.(2021)Song, Sohl-Dickstein, Kingma, Kumar, Ermon, and Poole]{song2021scorebased}
Yang Song, Jascha Sohl-Dickstein, Diederik~P Kingma, Abhishek Kumar, Stefano Ermon, and Ben Poole.
\newblock Score-based generative modeling through stochastic differential equations.
\newblock In \emph{ICLR}, 2021.

\bibitem[Song et~al.(2023)Song, Dhariwal, Chen, and Sutskever]{song2023consistency}
Yang Song, Prafulla Dhariwal, Mark Chen, and Ilya Sutskever.
\newblock Consistency models.
\newblock In \emph{Int. Conf. Mach. Learn.}, pages 32211--32252. PMLR, 2023.

\bibitem[Team et~al.(2024)Team, Ghosh, Walke, Pertsch, Black, Mees, Dasari, Hejna, Kreiman, Xu, Luo, Tan, Chen, Sanketi, Vuong, Xiao, Sadigh, Finn, and Levine]{octo2024octo}
Octo~Model Team, Dibya Ghosh, Homer Walke, Karl Pertsch, Kevin Black, Oier Mees, Sudeep Dasari, Joey Hejna, Tobias Kreiman, Charles Xu, Jianlan Luo, You~Liang Tan, Lawrence~Yunliang Chen, Pannag Sanketi, Quan Vuong, Ted Xiao, Dorsa Sadigh, Chelsea Finn, and Sergey Levine.
\newblock Octo: An open-source generalist robot policy, 2024.

\bibitem[Wang et~al.(2021)Wang, Qin, Zhang, and Fu]{wang2021neural}
Huan Wang, Can Qin, Yulun Zhang, and Yun Fu.
\newblock Neural pruning via growing regularization.
\newblock In \emph{ICLR}, 2021.

\bibitem[Wu et~al.(2024)Wu, Jing, Cheang, Chen, Xu, Li, Liu, Li, and Kong]{wu2024gr1}
Hongtao Wu, Ya Jing, Chilam Cheang, Guangzeng Chen, Jiafeng Xu, Xinghang Li, Minghuan Liu, Hang Li, and Tao Kong.
\newblock Unleashing large-scale video generative pre-training for visual robot manipulation.
\newblock In \emph{ICLR}, 2024.

\bibitem[Xie et~al.(2025)Xie, Chen, Zhao, YU, Zhu, Lin, Zhang, Li, Chen, Cai, Liu, Zhou, and Han]{xie2025sana1.5}
Enze Xie, Junsong Chen, Yuyang Zhao, Jincheng YU, Ligeng Zhu, Yujun Lin, Zhekai Zhang, Muyang Li, Junyu Chen, Han Cai, Bingchen Liu, Daquan Zhou, and Song Han.
\newblock {SANA} 1.5: Efficient scaling of training-time and inference-time compute in linear diffusion transformer.
\newblock In \emph{Int. Conf. Mach. Learn.}, 2025.

\bibitem[Xu et~al.(2024)Xu, Zhao, Xiao, and Hou]{xu2024ufogen}
Yanwu Xu, Yang Zhao, Zhisheng Xiao, and Tingbo Hou.
\newblock Ufogen: You forward once large scale text-to-image generation via diffusion gans.
\newblock In \emph{CVPR}, 2024.

\bibitem[Yue et~al.(2024)Yue, Wang, Kang, Han, Wang, Song, Feng, and Huang]{yue2024deervla}
Yang Yue, Yulin Wang, Bingyi Kang, Yizeng Han, Shenzhi Wang, Shiji Song, Jiashi Feng, and Gao Huang.
\newblock Deer-{VLA}: Dynamic inference of multimodal large language models for efficient robot execution.
\newblock In \emph{NeurIPS}, 2024.

\bibitem[Ze et~al.(2024)Ze, Zhang, Zhang, Hu, Wang, and Xu]{Ze2024DP3}
Yanjie Ze, Gu Zhang, Kangning Zhang, Chenyuan Hu, Muhan Wang, and Huazhe Xu.
\newblock 3d diffusion policy: Generalizable visuomotor policy learning via simple 3d representations.
\newblock In \emph{Proceedings of Robotics: Science and Systems (RSS)}, 2024.

\bibitem[Zhao et~al.(2023{\natexlab{a}})Zhao, Kumar, Levine, and Finn]{zhao23aloha}
Tony~Z. Zhao, Vikash Kumar, Sergey Levine, and Chelsea Finn.
\newblock Learning fine-grained bimanual manipulation with low-cost hardware.
\newblock In \emph{Robotics: Science and Systems}, 2023{\natexlab{a}}.

\bibitem[Zhao et~al.(2023{\natexlab{b}})Zhao, Xu, Xiao, and Hou]{zhaoMobileDiffusionSubsecondTextImage2023}
Yang Zhao, Yanwu Xu, Zhisheng Xiao, and Tingbo Hou.
\newblock Mobilediffusion: Subsecond text-to-image generation on mobile devices.
\newblock \emph{arXiv preprint arXiv:2311.16567}, 2023{\natexlab{b}}.

\bibitem[Zitkovich et~al.(2023)Zitkovich, Yu, Xu, Xu, Xiao, Xia, Wu, Wohlhart, Welker, Wahid, Vuong, Vanhoucke, Tran, Soricut, Singh, Singh, Sermanet, Sanketi, Salazar, Ryoo, Reymann, Rao, Pertsch, Mordatch, Michalewski, Lu, Levine, Lee, Lee, Leal, Kuang, Kalashnikov, Julian, Joshi, Irpan, Ichter, Hsu, Herzog, Hausman, Gopalakrishnan, Fu, Florence, Finn, Dubey, Driess, Ding, Choromanski, Chen, Chebotar, Carbajal, Brown, Brohan, Arenas, and Han]{zitkovich23rt2}
Brianna Zitkovich, Tianhe Yu, Sichun Xu, Peng Xu, Ted Xiao, Fei Xia, Jialin Wu, Paul Wohlhart, Stefan Welker, Ayzaan Wahid, Quan Vuong, Vincent Vanhoucke, Huong Tran, Radu Soricut, Anikait Singh, Jaspiar Singh, Pierre Sermanet, Pannag~R. Sanketi, Grecia Salazar, Michael~S. Ryoo, Krista Reymann, Kanishka Rao, Karl Pertsch, Igor Mordatch, Henryk Michalewski, Yao Lu, Sergey Levine, Lisa Lee, Tsang-Wei~Edward Lee, Isabel Leal, Yuheng Kuang, Dmitry Kalashnikov, Ryan Julian, Nikhil~J. Joshi, Alex Irpan, Brian Ichter, Jasmine Hsu, Alexander Herzog, Karol Hausman, Keerthana Gopalakrishnan, Chuyuan Fu, Pete Florence, Chelsea Finn, Kumar~Avinava Dubey, Danny Driess, Tianli Ding, Krzysztof~Marcin Choromanski, Xi Chen, Yevgen Chebotar, Justice Carbajal, Noah Brown, Anthony Brohan, Montserrat~Gonzalez Arenas, and Kehang Han.
\newblock Rt-2: Vision-language-action models transfer web knowledge to robotic control.
\newblock In \emph{Conference on Robot Learning}, pages 2165--2183. PMLR, 2023.

\end{thebibliography}
}

\clearpage
\setcounter{page}{1}
\renewcommand\thesection{\Alph{section}}
\renewcommand{\thefigure}{A\arabic{figure}}
\renewcommand{\thetable}{A\arabic{table}}
\setcounter{figure}{0} % Reset figure counter
\setcounter{table}{0}  % Reset table counter

\maketitlesupplementary

\section{Supplementary Material}
\label{sec:supplement}
% % 
% Having the supplementary compiled together with the main paper means that:
% % 
% \begin{itemize}
% \item The supplementary can back-reference sections of the main paper, for example, we can refer to \ref{sec:intro};
% \item The main paper can forward reference sub-sections within the supplementary explicitly (e.g. referring to a particular experiment); 
% \item When submitted to arXiv, the supplementary will already included at the end of the paper.
% \end{itemize}
% % 
% To split the supplementary pages from the main paper, you can use \href{https://support.apple.com/en-ca/guide/preview/prvw11793/mac#:~:text=Delete%20a%20page%20from%20a,or%20choose%20Edit%20%3E%20Delete).}{Preview (on macOS)}, \href{https://www.adobe.com/acrobat/how-to/delete-pages-from-pdf.html#:~:text=Choose%20%E2%80%9CTools%E2%80%9D%20%3E%20%E2%80%9COrganize,or%20pages%20from%20the%20file.}{Adobe Acrobat} (on all OSs), as well as \href{https://superuser.com/questions/517986/is-it-possible-to-delete-some-pages-of-a-pdf-document}{command line tools}.

The supplement consists of the following sections:
\begin{itemize}
  \item Section~\ref{sec:supp:extensive_exp_dpt} presents the extensive experimental results on Robomimic dataset based on the DiffusionPolicy Transformer (DP-T) model.
  \item Section~\ref{sec:supp:real-world_exp} describes the real-world experiments based on DP-T and MoDE models, including the experimental setup and results.
\end{itemize}
We provide a webpage to visualize the results of the pruned models and original models, which can be found at \url{https://weleen.github.io/LightDP/}.

\section{Extensive Experiments based on DP-T}
\label{sec:supp:extensive_exp_dpt}
\begin{table}[htbp]
\centering

\resizebox{0.47\textwidth}{!}{
  \begin{tabular}{lcccccc}
    \toprule
    \textbf{Models} & \textbf{Lift-ph} & \textbf{Can-ph} & \textbf{Square-ph} & \textbf{Transport-ph} & \textbf{Push-T} & \textbf{ToolHang-ph} \\
    \midrule
    DP-T        & 1.000 & 1.000 & 1.000 & 0.955 & 0.772 & 0.713 \\
    DP-T-D6/6-8  & 1.000 & 1.000 & 1.000 & 0.950 & 0.752 & 0.707 \\
    \midrule
    \textbf{Models} & \textbf{Lift-mh} & \textbf{Can-mh} & \textbf{Square-mh} & \textbf{Transport-mh} & \textbf{Kitchen} & \textbf{Block Push}\\
    \midrule
    DP-T        & 1.000 & 1.000 & 0.940 & 0.727 & 0.574 & 1.000\\
    DP-T-D6/6-8  & 1.000 & 1.000 & 0.955 & 0.773 & 0.571 & 1.000\\
    \bottomrule
  \end{tabular}
}
\caption{The extensive evaluation on DP-T tasks (Push-T and Robomimic), showing the success rates of the original model (DP-T) and the pruned model (DP-T-D6/6-8). The pruned model maintains performance across most tasks, with only minor drops in success rates.}
\label{tab:extensive_exp_dpt}
\end{table}

In Table~\ref{tab:extensive_exp_dpt}, we have provided success rate on all tasks (\ie, Push-T and Robomimic) in the Diffusion Policy~\cite{chi2023diffusion} work, which indicate that the pruned model DP-T-D6/6-8 preserves the baseline's performance on most tasks, and the performance only drops by less than 0.02 on the tasks.

\section{Real-world Experiments}
\label{sec:supp:real-world_exp}
\begin{figure}[t]
    \centering
    \animategraphics[autoplay,loop,nomouse,poster=first,width=\linewidth]{8}{figures/frames/}{01}{83}
    \caption{Real-world experiments for DP-T (first column) and MoDE (other columns). Task descriptions are shown below each image. \textcolor{red}{This figure contains an animated video. For optimal viewing, please zoom in and use a professional PDF reader.}}
    \label{fig:real_world_exp}
\end{figure}
\begin{table}[htbp]
  \centering
  \begin{subtable}[t]{0.16\textwidth}
    \centering
    \resizebox{\textwidth}{!}{
      \begin{tabular}{lc}
        \toprule
        \textbf{Models} & \textbf{Task 1} \\
        \midrule
        DP-T         & 0.80 \\
        DP-T-D6/6-8   & 0.75 \\
        \bottomrule
      \end{tabular}
    }
    % \caption{}
    \label{tab:supp_real_world_success_inovo}
  \end{subtable}
  \begin{subtable}[t]{0.3\textwidth}
    \resizebox{\textwidth}{!}{
      \begin{tabular}{lccc}
        \toprule
        \textbf{Models} & \textbf{Task 2} & \textbf{Task 3} & \textbf{Task 4} \\
        \midrule
        MoDE            & 0.80 & 0.55 & 0.30 \\
        MoDE-10/10-12   & 0.75 & 0.50 & 0.30 \\
        \bottomrule
      \end{tabular}
    }
    % \caption{}
    \label{tab:supp_real_world_success_lebai}
    \end{subtable}
  \caption{Real-world evaluation results based on DP-T (on a Inovo Robot) and MoDE (on a Lebai Robot). The success rates are shown for each task, with the pruned model (DP-T-D6/6-8 and MoDE-10/10-12) maintaining performance across most tasks, with only minor drops in success rates.}
% \vspace{-1.5em}
\label{tab:supp_real_world_success}
\end{table}

Based on two models DP-T and MoDE, we deploy our \emph{LightDP} on two robotic arms (an Inovo robot for DP-T and a Lebai robot for MoDE), where each task is executed by 20 times. As shown in Figure~\ref{fig:real_world_exp} and Table~\ref{tab:supp_real_world_success}, the pruned model achieves a comparable success rate on these real-world tasks. Considering that most household users are often redundant to purchase advanced device, we selected the most accessible and portable device (\ie, iPhone) as the computing platform for our robotic development setup. Moreover, we also evaluate our approach based on a Jetson Orin NX (16 GB, Jetpack 5.1.1), the latency is 244.68ms (\textit{resp.}, 37.69ms) based on DP-T (\textit{resp.}, DP-T-D6/6-8).

\end{document}